%% file: main.tex
\documentclass[pdflatex,sn-mathphys-num]{sn-jnl}% Math and Physical Sciences Numbered Reference Style 
\usepackage{graphicx}%
\usepackage{multirow}%
\usepackage{amsmath,amssymb,amsfonts}%
\usepackage{amsthm}%
\usepackage{mathrsfs}%
\usepackage[title]{appendix}%
\usepackage{xcolor}%
\usepackage{textcomp}%
\usepackage{manyfoot}%
\usepackage{booktabs}%
\usepackage{algorithm}%
\usepackage{algorithmicx}%
\usepackage{algpseudocode}%
\usepackage{listings}%
\usepackage{enumitem}

\usepackage{acronym}  % for acronym
\usepackage{import}
\usepackage{tcolorbox}
\theoremstyle{thmstyleone}%
\theoremstyle{thmstyletwo}%

\theoremstyle{thmstylethree}%
\newtheorem{definition}{Definition}%

\begin{document}

\title[Safe and Robust Reinforcement Learning]{Safe and Robust Reinforcement Learning: Principles and Practice}

\author*[1]{\fnm{Taku} \sur{Yamagata}}\email{taku.yamagata@bristol.ac.uk}

\author[1]{\fnm{Ra{\'u}l} \sur{Santos-Rodr{\'i}guez}}\email{ensr@bristol.ac.uk}

\affil*[1]{\orgdiv{Intelligent System Laboratory}, \orgname{University of Bristol}, \orgaddress{\city{Bristol}, \country{UK}}}

\abstract{Reinforcement Learning (RL) has shown remarkable success in solving relatively complex tasks, yet the deployment of RL systems in real-world scenarios poses significant challenges related to safety and robustness. This paper aims to identify and further understand those challenges thorough the exploration of the main dimensions of the safe and robust RL landscape, encompassing algorithmic, ethical, and practical considerations. We conduct a comprehensive review of methodologies and open problems that summarizes the efforts in recent years to address the inherent risks associated with RL applications.

After discussing and proposing definitions for both safe and robust RL, the paper categorizes existing research works into different algorithmic approaches that enhance the safety and robustness of RL agents. We examine techniques such as uncertainty estimation, optimisation methodologies, exploration-exploitation trade-offs, and adversarial training. Environmental factors, including sim-to-real transfer and domain adaptation, are also scrutinized to understand how RL systems can adapt to diverse and dynamic surroundings. Moreover, human involvement is an integral ingredient of the analysis, acknowledging the broad set of roles that humans can take in this context. 

Importantly, to aid practitioners in navigating the complexities of safe and robust RL implementation, this paper introduces a practical checklist derived from the synthesized literature. The checklist encompasses critical aspects of algorithm design, training environment considerations, and ethical guidelines. It will serve as a resource for developers and policymakers alike to ensure the responsible deployment of RL systems in many application domains.}

\keywords{reinforcement learning, safe AI, robust Markov decision process, constrained Markov decision process}

\maketitle

\section{Structure}\label{intro}

This paper provides a general overview of the definitions, approaches and practical considerations for safe and robust \ac{RL}. We intend to cover a wide range of topics and approaches related to safe and robust \ac{RL}, but fully accepting that these are not only ambiguously defined and used, but fast evolving concepts.

Here, we start by introducing a basic formulation of \ac{RL} framework in Sec.~\ref{sec01:rl}. We collect the most common definitions of the terms -- \textit{safety} and \textit{robustness} in Sec.~\ref{sec02:preliminaries}. Based on those, we propose two working definitions for the remainder of the work. 

\begin{figure}[H]
    \centering
    \includegraphics[width=1\textwidth]{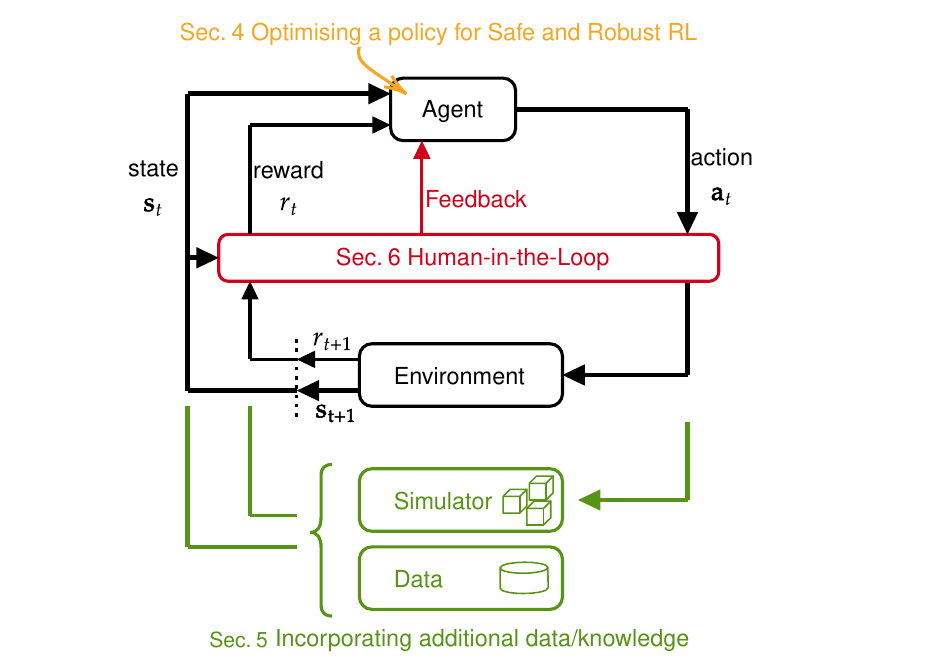}
    \caption{Overview of the reinforcement learning setting (in black) with relevant components (coloured) to the safe and robust RL introduced in Sec.\ref{sec02:optimisation} to Sec.\ref{sec02:human-in-the-loop}}
    \label{fig02:overviewRL_all}
\end{figure}

The following three sections (Sec.~\ref{sec02:optimisation} to \ref{sec02:human-in-the-loop}) introduce and categorize various safe and robust RL approaches.
Figure~\ref{fig02:overviewRL_all} shows the overview of \ac{RL} framework and highlights where the relevant categories introduced in these sections fit. We provide a summary of the literature introduced in this paper in Appendix~\ref{literature_summary}. It includes two figures (Fig.~\ref{fig02:literature_safe} and \ref{fig02:literature_robust}) that show an overview of both the safe \ac{RL} and robust \ac{RL} literature, placing the works in chronological order (from top to bottom). 

Section~\ref{sec02:optimisation} focuses on how to train the agent's policy to achieve safety and robustness. It has three main components -- criteria, method and exploration. The \texttt{optimisation criteria} related to the study of objective functions that are used to achieve safety. The \texttt{optimisation method} provides a overview of approaches to achieve the criteria. The \texttt{exploration} part focuses on methods for exploration that count with a safety ingredient. Exploration is a key research topic because exploration and safety are conflicting concepts -- in practice, it is a hard trade-off to explore while maintaining the safety.

Section~\ref{sec02:external_knowledge} discusses approaches incorporating additional data/knowledge -- data, simulators and human knowledge. 
In principle, collecting more information about the environment, the more likely to be able to improve on the safety front. We explore the different solutions for adding these extra sources of information into the learning process, while assessing its value. Section~\ref{sec02:human-in-the-loop} deals specifically with \textit{human-in-the-Loop}, which usually sits in between the agent and environment, intervening, interfacing and guiding the interactions. For example, we present alternatives for humans to give feedback or \textit{shaping} the reward, or humans changing the agent's action altogether to maintain safety. 

In subsequent sections (Sec.~\ref{sec02:related_topics} and \ref{sec02:ethical_issues}), we explore a broad spectrum of topics relevant to different aspects of safety and robustness of \ac{RL}. For example, until this point, the discussion is centered around the standard RL framework. Section~\ref{sec02:related_topics} looks into alternative \ac{RL} paradigms, specifically multi-agent \ac{RL} and hierarchical \ac{RL}, as well as other domains that are intrinsically linked to the concept of safe and robust \ac{RL}. In Sec.~\ref{sec02:ethical_issues}, we discuss ethical aspects of \ac{RL} agents for real-world safety critical applications, as we argue that the definition of ethical requirements tailored to the target application domain should be prevalent in any development and deployment. 

In Sec.~\ref{sec02:check_list}, we design and provide practitioners with a checklist for designing a safe and robust \ac{RL} system. It summarises important design choices and offers a workflow to develop and deploy a safe and robust RL solutions in practice.

Each section of this paper is designed to be stand-alone, requiring no prior knowledge from other sections. However, understanding a basic \ac{RL} formulation, as outlined in Section~\ref{sec01:rl}, is assumed. Readers may directly navigate to sections that align with their interests. 

\section{Reinforcement Learning}
\label{sec01:rl}

\begin{figure}[h!]
	\centering
	\includegraphics[width=1\textwidth]{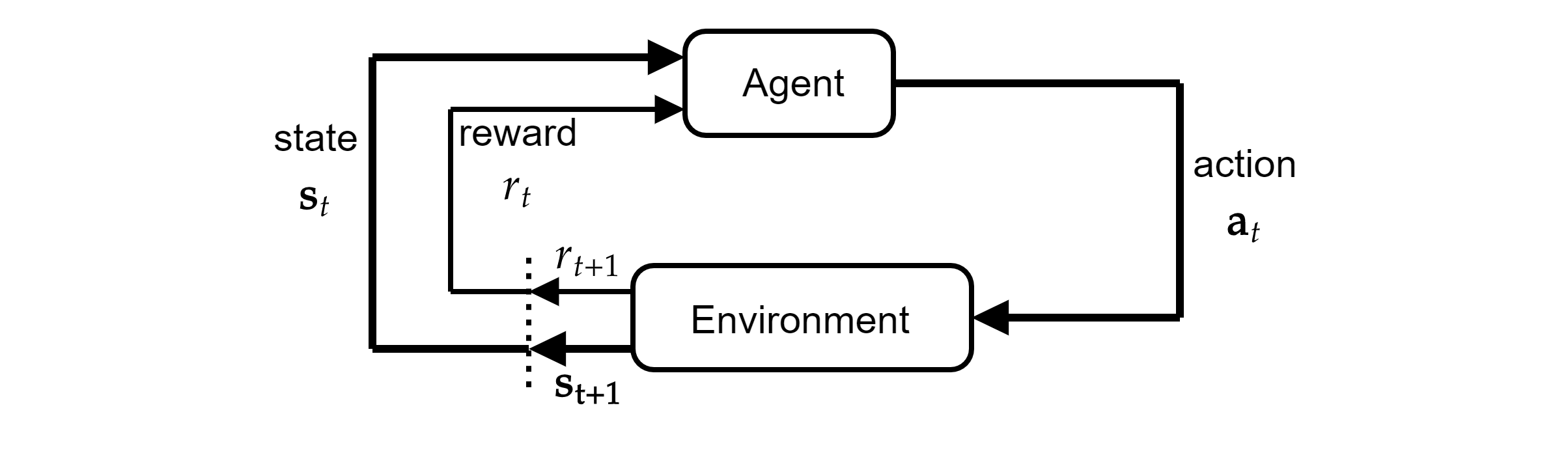}
	\caption[Overview of the reinforcement learning setting]{Overview of the reinforcement learning setting.}
	\label{fig01:overviewRL}
\end{figure}

Here, we introduce a minimal formulation of \Ac{RL} paradigm. \Ac{RL} is an abstract framework for any learning process that involves sequentially interacting with an environment to achieve a certain objective~\cite{Sutton1998}. The learner is called the \textit{agent}. It interacts with the \textit{environment}, observes its consequences, and receives a reward (or a cost) signal -- a special numerical assessment the current situation. The agent outputs a sequence of actions to maximise the cummulative reward (or minimise the cost) as shown in Fig.~\ref{fig01:overviewRL}.

More formally, the agent and environment interactions are discretised into a sequence of time steps, $t=0, 1, 2, \dots$. At each time step $t$, the agent observes the state of the environment $\mathbf{s}_t$, then decides an action $\mathbf{a}_t$. In the next time step, the environment updates the state based on the agent's action -- it becomes $\mathbf{s}_{t+1}$, and also generates the reward $r_{t+1} \in \mathbb{R}$.
The agent basically learns a mapping from the state to the action that maximises the total amount of reward it receives over the long run. The mapping is called \textit{policy} denoted as $\pi_t\left(\mathbf{a}|\mathbf{s}\right)$ that indicates the probability of $\mathbf{a}_t = \mathbf{a}$ when the state is $\mathbf{s}_t = \mathbf{s}$.

Reinforcement learning differs from traditional \textit{supervised learning}. While the latter learns from examples provided by experts, \ac{RL} learns from its interactions with the environment. \ac{RL} is suitable when it is obvious what we would like to achieve (the goal), but it is not so obvious how to achieve it. For example, it is clear what we would like to achieve in the chess game (taking the opponent's King and winning), but it is not apparent how to achieve it. In general, \ac{RL} learns how to achieve the goal autonomously from the interactions (playing games) via a trial-and-error approach.

% ========================================================================================
% Chapter 2 BEGIN
% ========================================================================================
\section{Definitions}
\label{sec02:preliminaries}

In this Section, instead of starting by providing the reader with a formal definition of safe and robust RL, we collect and discuss definitions already present in the literature for each concept. We aim to show the diversity and lack of consesus, while focusing on the similarities among definitions. 

\subsection{Robust reinforcement learning}
\label{sec02:robustRL_definitions}

According to the Cambridge dictionary\footnote{https://dictionary.cambridge.org/dictionary/english/robust}, the definitions of the word \textit{robust} refer to ``(of a person or animal) strong and healthy, or (of an object or system) strong and unlikely to break or fail.'' Also, according to Wikipedia\footnote{https://en.wikipedia.org/wiki/Robustness}, \textit{robustness} ``is the property of being strong and healthy in constitution. When it is transposed into a system, it refers to the ability of tolerating perturbations that might affect the system’s functional body. In the same line robustness can be defined as the ability of a system to resist change without adapting its initial stable configuration.''

Certain uses of \textit{robust \ac{RL}} are relatively consistent across the all \ac{RL} literature. To the best of our knowledge, the term \textit{robust \ac{RL}} is first introduced by Jun Morimoto and Kenji Doya~\cite{morimoto2005robust}. They define it as,
\begin{quote}
    ``a new reinforcement learning (RL) paradigm that explicitly takes into account input disturbance as well as modeling errors.''
\end{quote}
Concurrently, G. Iyengar also proposed robust dynamic programming~\cite{iyengar2005robust}, a method aimed to achieve robust \ac{RL}. It defines the method as:
\begin{quote}
    ``Systematically mitigate the sensitivity of the dynamic programming optimal policy to ambiguity in the underlying transition probabilities''
\end{quote}
More recent survey papers suggest the following definitions, which align with the above Moriomoto and Doya's definition. 
\begin{quote}
    ``Robust RL aims to learn a robust optimal policy that accounts for model uncertainty of the transition probability to systematically mitigate the sensitivity of the optimal policy in perturbed environments''~\cite{chen2020overview} % An Overview of Robust Reinforcement Learning
\end{quote}
\begin{quote}
    ``robustness—in the scope considered in this survey—refers to the ability to cope with variations or uncertainty of one’s environment. In the context of reinforcement learning and control, robustness is pursued w.r.t. specific uncertainties in system dynamics, e.g., varying physical parameters''~\cite{moos2022robust} % Robust Reinforcement Learning: A Review of Foundations and Recent Advances
\end{quote}

In this paper, we slightly broaden the definition and consider uncertainty in all information the agent receives (not just uncertainty in the environment). We propose then to consider \textit{robust RL} as follows.

\begin{definition}[Robust RL]
Robust RL methodologies are those than can cope with (or systematically mitigate the sensitivity of) all the relevant sources of uncertainty of the environment, taking into account any other information that the agent receives.
\end{definition}

\subsection{Safe reinforcement learning}
\label{sec02:safeRL_definitions}

Following a similar procedure, according to the Cambridge dictionary\footnote{https://dictionary.cambridge.org/dictionary/english/safe}, the general definition of the word \textit{safe} is ``not in danger or likely to be harmed''. Also according to Wikipedia\footnote{https://en.wikipedia.org/wiki/Safety}, the word \textit{safety} is defined as 
``the state of being safe, the condition of being protected from harm or other danger. Safety can also refer to the control of recognized hazards in order to achieve an acceptable level of risk.''

While the concept of \textit{safety} in \ac{RL} is relatively clear, the terminology \textit{safe \ac{RL}} is heavily overloaded. Roughly the following four aspects of safe \ac{RL} are considered in the literature. 
\begin{enumerate}[label=\Roman*.]
	\item Consistent performance. 
	\item Maintaining safety constraints. 
	\item Aligned with the true objective.
	\item Accepting human intervention.
\end{enumerate}

\subparagraph{I) Consistent performance} requires the agent to always perform well in various conditions. This requirement is similar to the robust \ac{RL}. 
\subparagraph{II) Maintaining safety constraints} requires the agent to maintain certain constraints defined by the system. \subparagraph{III) Aligned with the true objective} means that the agent's objective must be aligned with the task's true objective (or human intention). The agent cannot be safe if the given objective is not aligned with the true objective, no matter how good the agent algorithm is. 
\subparagraph{IV) Accepting human intervention} indicates that the agent (or system) must have a mechanism for human intervention (or ``emergency stop''). We do not think this feature alone makes the agent safe; rather, this feature is mandatory for all real-world \ac{RL} applications. 

Below, we introduce several definitions of safe RL in literature with the Roman numbers (I to IV) that indicate which category the definition belongs.

The most generic (and high level) definition of \textit{safe RL} would be by Javier Garc{\i}a et al.~\cite{garcia2015comprehensive}
\begin{quote}
    ``the process of learning policies that maximize the expectation of the return in problems in which it is important to ensure reasonable system performance and/or respect safety constraints during the learning and/or deployment processes'' I),II) % A Comprehensive Survey on Safe Reinforcement Learning
\end{quote}

The definition by Shangding Gu et al.~\cite{gu2022review} explicitly includes adversary attacks while accommodating a broader sense of \textit{safety}, which includes mitigating undesirable situations and reducing risk.
\begin{quote}
    ``about optimizing cost objectives, avoiding adversary attacks, improving undesirable situations, reducing risk, and controlling agents to be safe'' II) % "A Review of Safe Reinforcement Learning: Methods"
\end{quote}

Yongshuai Liu et.al.~\cite{liu2021policy} and Lukas Brunke et al.~\cite{brunke2022safe} employ the concept of cost and define \textit{safety} as controlling the cost. This definition is originated from a \ac{CMDP}~\cite{altman1999constrained} framework.
\begin{quote}
    ``The safe RL agent's objective is to maximise long-term reward while keeping certain costs under their respective constraints.'' II) % "Policy Learning with Constraints in Model-free RL: A Survey" and "Safe Learning in Robotics: From Learning-Based Control to Safe RL"
\end{quote}

Dylan Hadfield-Menell et al.~\cite{hadfield2016cooperative} argue from slightly different prospective and define the \textit{safe RL} as:
\begin{quote}
    ``Safe RL has a mechanism for a human to interfere the agent effectively.'' IV)  %"Cooperative Inverse Reinforcement Learning"
\end{quote}

This definition relates to the following note by Norbert Wiener~\cite{wiener1960some} in one of the earliest explanations of the problems that arise when a powerful autonomous system operates with an incorrect objective.
\begin{quote}
    “If we use, to achieve our purposes, a mechanical agency with whose operation we cannot interfere effectively . . . we had better be quite sure that the purpose put into the machine is the purpose which we really desire.” III) or IV)
\end{quote}

Wiener's notion could lead to two lines of approaches to the safety of the autonomous system. One introduces an effective intervention mechanism, and the other has the right reward function, and an algorithm can achieve the goal -- indeed, most \textit{safe RL} approaches are belong to either of these lines.

In this paper, we define \textit{safe RL} in the similar way as Javier Garc{\i}a et al.~\cite{garcia2015comprehensive}, but also added aspects of III) and IV) as Dylan Hadfield-Menell et al.~\cite{hadfield2016cooperative}.

\begin{definition}[Safe RL]
    {Safe RL is the process of learning policies that maximize the expectation of the return in problems in which it is important to ensure reasonable system performance and/or respect safety constraints during the learning and/or deployment processes. Also, the system must have the right objectives (the reward function aligned with the objective of the task) and a mechanism for humans to intervene.}  
\end{definition}
This definition also covers the robust \ac{RL} aspect as well, so we use it as our working definition of safe and robust \ac{RL} for the remainder of the paper.

\section{Optimising a policy for Safe and Robust RL}
\label{sec02:optimisation}
This section outlines various training methodologies essential for developing a safe and robust \ac{RL} agent. 
Initially, we present helpful concepts such as \ac{MDP}, robust \ac{MDP}, and constrained \ac{MDP}, which are foundational for understanding different optimisation strategies. Subsequently, we explore diverse optimisation criteria and techniques to fulfil these criteria to ensure a safe and robust \ac{RL}.

\subsection{Robust and Constrained Markov Decision Process}
\ac{RL} framework can be seen as a Markov decision process when the environment and the agent hold the Markovian property. It can be extended to a robust Markov process and a constrained Markov process. They are closely related to robust and safe \ac{RL}, and they are very useful concepts to define some of safe and robust \ac{RL} algorithms, so this section describes the definitions of these Markov processes.

A \acf{MDP} is defined as a tuple $(\mathcal{S},\mathcal{A},r,\mathcal{P},\mu,\gamma)$~\cite{Sutton1998}, where $\mathcal{S}$ is the set of states, $\mathcal{A}$ is the set of actions, $r : \mathcal{S}\times \mathcal{A}\times \mathcal{S} \mapsto \mathbb{R}$ is the reward function, $\mathcal{P} : \mathcal{S}\times \mathcal{A}\times \mathcal{S} \mapsto \left[0,1\right]$ is the state transition probability function, $\mu : \mathcal{S} \mapsto \left[0,1\right]$ is the initial state distribution and $\gamma \in \left[0,1\right]$ is the discount factor for the future reward. A policy $\pi : \mathcal{S} \mapsto P(\mathcal{A})$ is a mapping from states to a probability distribution over actions. A standard \ac{MDP} aims to learn a policy $\pi$ that maximises the discounted cumulative reward:
\begin{equation} \label{ch2_mdp_objective}
	\arg\max_{\pi} J_r^{\pi} = \mathbb{E}_{\tau\sim\pi}\left[ \sum_{t=0}^{\infty}\gamma^t r\left(s_t, a_t, s_{t+1}\right)\right],
\end{equation}
where $\tau=(s_0, a_0, s_1, a_1, \cdots)$ denotes a trajectory, and $\tau\sim\pi$ denotes trajectories sampled from the policy $\pi$. 
%In some applications, it is more natural to employ a cost function $c : \mathcal{S}\times \mathcal{A} \times \mathcal{S} \mapsto \mathbb{R}$ instead of the reward function $r$ and the aim of \ac{MDP} becomes finding a policy $\pi$ that minimises the discounted accumulative cost.

An \ac{RMDP} extends the definition of the standard \ac{MDP} by introducing $\mathscr{P}$ an uncertainty set for the state transition probabilities. An \ac{RMDP} is defined as a tuple $(\mathcal{S}, \mathcal{A}, r, \mathscr{P}, \mu,\gamma)$. It guarantee the highest discounted accumulative reward with the given uncertainty set:
\begin{equation} \label{ch2_rmdp_objective}
	\arg\max_{\pi} J_{r,\mathscr{P}}^{\pi} = \inf_{\mathcal{P} \in \mathscr{P}} \mathbb{E}_{\tau\sim\pi, \mathcal{P}}\left[ \sum_{t=0}^{\infty}\gamma^t r\left(s_t, a_t, s_{t+1}\right)\right],
\end{equation}
where $\tau\sim\pi, \mathcal{P}$ denotes sampled trajectories from the policy $\pi$ and the state transition probability $\mathcal{P}$. 
The uncertainty set $\mathscr{P}$ is typically for the state transition probabilities. However, in general, it can be for any parameters in the target environment.

A \acf{CMDP} is also a concept that extends the standard \ac{MDP} by introducing cost functions $C$ in addition to the reward function. It is defined as a tuple $(\mathcal{S},\mathcal{A}, r, C, \mathcal{P},\mu,\gamma)$. The cost functions $c_i \in C, c_i : \mathcal{S}\times \mathcal{A}\times \mathcal{S} \mapsto \mathbb{R}$ are constrained suitably for the target application. The types of constraints are discussed in the following section. A \ac{CMDP} aims to learn a policy $\pi$ that maximises the discounted cumulative reward while it satisfies all of its necessary constraints. Formally, the \ac{CMDP} becomes the following conditional optimisation problem:
\begin{equation} \label{ch2_cmdp_objective}
\begin{split}
    \arg\max_{\pi} J_r^{\pi} &= \mathbb{E}_{\tau\sim\pi}\left[ \sum_{t=0}^{\infty}\gamma^t r\left(s_t, a_t, s_{t+1}\right)\right], \\
    &s.t.\ J_{c_i}^\pi \leq \epsilon_i \forall i,
\end{split}
\end{equation}
where $J_{c_i}^\pi$ denotes a statistical measure over the i-th cost function values from the trajectory sampled with the policy $\pi$ and $\epsilon_i$ is the allowed upper bound for the measure. Various types of constraints are realised by defining appropriate $J_{c_i}^\pi$.

\subsection{Optimisation criteria}
\label{sec02:optimisation_criteria}

In this subsection, we will review two optimisation criteria commonly used for safe and robust RL. 
For a normal \ac{RL} setting, the optimisation criterion is to maximise the total (discounted) reward. A safe and robust RL scenario requires slightly different criteria. 
The first type is robust RL criteria, which aims to maximise the expected total reward under some worst-case scenarios or distributional assumptions. The second type is constrained RL criteria, which imposes additional constraints on the agent's actions or outcomes. We will describe these criteria in the rest of this section, and discuss some approaches to achieve them in the following section.

\paragraph{Robust RL criterion} The former optimisation criterion indicates that the agent consistently achieves a certain level of accumulated reward within a given uncertainty in the \ac{RL} framework. Often, it assumes a certain level of uncertainty in the environment and maximises the expected total reward in the worst case. This category is equivalent to the robust \ac{RL}~\cite{morimoto2005robust}. This task setup is also often referred to \ac{RMDP}~\cite{iyengar2005robust}.

\begin{figure}[h!]
	\centering
	\includegraphics[width=1\textwidth]{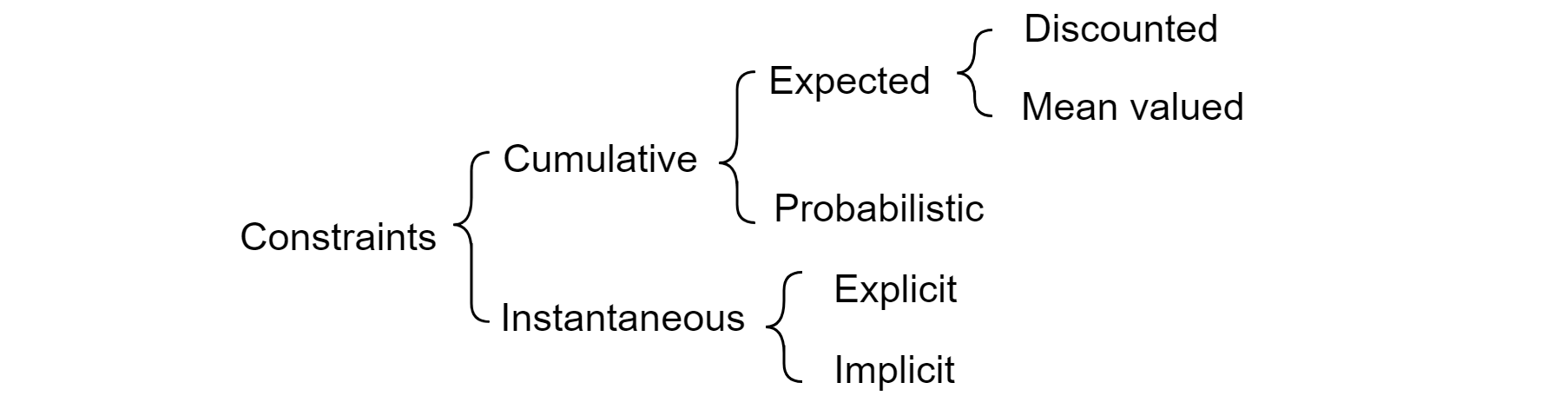}
	\caption[Categories of constraints]{Categories of constraints.}
	\label{fig02:constraints}
\end{figure}

\paragraph{Constrained RL criterion} The latter definition is more common in the recent safe RL literature. Typically it introduces a cost function in addition to the reward function, and the agent tries to maximise the expected accumulated reward while maintaining particular constraints regarding the cost. The constraints can be categorised as Fig.\ref{fig02:constraints}~\cite{liu2021policy}.
The cumulative constraints require the sum or mean of a given cost to be within a specific limit. e.g. for an electric car navigation task, the sum of energy consumption (cost) must be less than the available battery (limit) before reaching a charging station (or home). The cumulative constraints are split into two categories -- expectations and probability of the costs. The former includes discounted cumulative constraints and mean-valued constraints~\cite{altman1999constrained}.

An expected discounted cumulative constraint is of the form
\begin{equation} \label{ch2_discount_constraint}
J_{c_i}^{\pi} = \mathbb{E}_{\tau\sim\pi}\left[ \sum_{t=0}^{\infty} \gamma^t c_i\left(s_t, a_t, s_{t+1}\right)\right] \leq \epsilon_i.
\end{equation}

An expected mean valued cumulative constraint is of the form
\begin{equation} \label{ch2_mean_constraint}
J_{c_i}^{\pi} = \mathbb{E}_{\tau\sim\pi}\left[ \frac{1}{T}\sum_{t=0}^{T-1} c_i\left(s_t, a_t, s_{t+1}\right)\right] \leq \epsilon_i.
\end{equation}
While the probabilistic constraints~\cite{geibel2006reinforcement} require the probability of the cumulative costs exceeding a threshold ($\eta$) is less than $\epsilon_i$.
\begin{equation} \label{ch2_probabilistic_constraint}
J_{c_i}^{\pi} = P\left(\sum_{t} c_i\left(s_t, a_t, s_{t+1}\right) > \eta \right) \leq \epsilon_i.
\end{equation}
Instantaneous constraints are constraints on the actions, states or cost functions that must satisfy for each time step. They can be further split into explicit and implicit cases. An explicit constraint has a closed-form expression that can be numerically checked. On the other hand, an implicit constraint does not have an accurate closed-form formulation due to the complexity of the system. Hence it requires learning a function that checks if a given state action pair satisfy requirements.  The instantaneous constraints form as
\begin{equation} \label{ch2_expricit_constraint}
J_{c_i}^{\pi} = c_i\left(s_t, a_t, s_{t+1}\right)\leq \epsilon_i.
\end{equation}
It is hard to show a generic form of implicit constraints. However, instantaneous probabilistic constraints are one of the implicit constraints and can be form of
\begin{equation} \label{ch2_implicit_constraint}
J_{c_i}^{\pi} = P\left(c_i\left(s_t, a_t, s_{t+1}\right) > \eta \right) \leq \epsilon_i.
\end{equation}

\subsection{Optimisation methods}
\label{sec02:optimisation_methods}

Here, we explore strategies to achieve the criteria introduced in the previous section. We begin by considering optimisation methods for exploitation policy, which solely aim to achieve the optimisation criteria introduced in the above subsection. Subsequently, we introduce methods for safe exploration, which aim to gather diverse training data for training the exploitation policy while maintaining safety. 

\paragraph{Optimising exploitation policy for robust RL criterion}
For the robust \ac{RL} criterion, the most straightforward approach would be maximising the reward in the worst case. There are three lines of approaches to achieving empirically robust performance -- i) robust adversarial approaches, ii) domain randomisation approaches and iii) approaches employing a statistical measure.

i) The robust adversarial approaches combine \ac{RL} with adversarial learning that learns a policy to control the environment's parameters to minimise the reward that \ac{RL} agent obtains while the \ac{RL} learns a policy maximises it~\cite{pinto2017robust, pan2019risk}. These two policies are updated alternately (each is updated while fixing the other), attempting to progressively improve the robustness of the \ac{RL} agent's policy and the strength of its adversary. The difficulty of the approach is that it requires a balance between the ability of \ac{RL} agent and its adversarial agent. If the adversarial agent possesses too much control ability, \ac{RL} agent fails to learn any helpful policy, while too little control results in not robust \ac{RL} policy.

ii) The domain randomisation approaches learn a policy that empirically generalises to a broader range of tasks or environments. Instead of considering the worst-case, it learns from an environment with randomly perturbated parameters~\cite{sadeghi2016cad2rl, loquercio2019deep}. These parameters often have pre-specified ranges.  

iii) The approaches employing a statistical measure. Some of the approaches for maximising the worst case scenario may end up with a too-pessimistic solution and performs poorly in the typical condition. An alternative approach would consider a soft worse case, which employs a statistical metric instead of the worst reward such as CVaR~\cite{rockafellar2000optimization} or certain percentile point. These statistical measures can be obtained by employing distributional reinforcement learning approaches~\cite{bellemare2017distributional,dabney2018distributional,dabney2018implicit} (such as \ac{IQN}~\cite{dabney2018implicit}) that learn the distribution of the value function. However, \ac{IQN} alone does not capture the Epistemic uncertainties, which is the uncertainty due to lack of knowledge. Epistemic uncertainty can be captured by employing an ensemble approach. This approach is proposed as \ac{EQN}~\cite{hoel2023ensemble} that combines \ac{IQN} and the ensemble approach to capture Epistemic and Aleatoric uncertainties and learn an appropriate value function distribution for computing the statistical measures like CVaR.
 
\paragraph{Optimising exploitation policy for constrained RL criterion}
For the constrained \ac{RL} criterion, Lagrangian relaxation is the most straightforward approach to address the constraints~\cite{altman1999constrained,chow2015risk,tessler2018reward}. 
The general form of Lagrangian relaxation is to reduce the problem to an unconstrained problem with Lagrange multipliers. 
These adaptive Lagrange multipliers are then used to penalise constraint violations as
\begin{equation} \label{ch2_lagrangian_method}
\min_{\lambda_i\geq0}\max_\pi L(\pi, \lambda)=J_r^\pi-\sum_i \lambda_i\left(J_{c_i}^\pi-\epsilon_i\right).
\end{equation}
However, this approach is sensitive to the initialisation of the Lagrange multipliers and the learning rate. Moreover, the Lagrangian multipliers are solved on a slower time scale, making it difficult to optimise in practice~\cite{achiam2017constrained, chow2017risk}.
Another alternative approach employs a particular function to incorporate the constraints and merge them into a single objective without Lagrange multipliers. One such example is \ac{IPO}~\cite{liu2020ipo}, a first-order constrained method inspired by the interior-point method~\cite{boyd2004convex}. \ac{IPO} employs logarithmic barrier functions as penalty functions to accommodate the constraints as Eq.~\ref{ch2_ipo}. It shows the $\log$ term go to minus infinity as $J_{c_i}^\pi$ getting closer to $\epsilon_i$.
\begin{equation} \label{ch2_ipo}
\max_\pi J_r^{\pi} + \sum_i \frac{1}{t_i} \log\left(-J_{c_i}^\pi+\epsilon_i\right),
\end{equation}
where $t_i$ is a hyperparamter. 

For methods based on strong theoretical justifications, a line of works~\cite{kakade2002approximately, pirotta2013safe} guarantees performance improvement for each policy update. \Ac{TRPO}~\cite{schulman2015trust} adopted it to \ac{DNN} parameterised polices, and its successor \ac{PPO}~\cite{schulman2017proximal} established better empirical performances with a much simpler algorithm. \Ac{CPO}~\cite{achiam2017constrained} is inspired by \ac{TRPO}, and it guarantees keeping the constraints and performance improvement for each iteration. However, \ac{CPO} is computationally expensive as it uses conjugate gradients for approximating the Fisher Information Matrix, whose approximation error affects the overall performance. Furthermore, \ac{CPO} only supports constraints that satisfy the Recursive Bellman Equation (i.e. discounted sum constraints), and it is difficult to accommodate multiple constraints.

\paragraph{Exploration policy}
\label{sec02:sefe_exploration}
In the standard \ac{RL} setting, the training dataset, or trajectory, is generated through interaction with the environment. This means that the quality of the training data is determined by the policy guiding these interactions. In essence, the exploration policy has a direct impact on the quality of data available for training the agent. It's critical that the policy not only seeks out new states and actions to enrich the training data but also ensures safety. As further elaborated in this paragraph, striking a delicate balance between safety and exploration is essential.

Common \ac{RL} exploration strategies often rely on some stochasticity in the action choices, such as the $\epsilon$-greedy algorithm. Although they are simple and yet effective in many \ac{RL} tasks, it does not consider any risk. Hence it might cause many catastrophic failures during the learning process. Moreover, it may result in constant failures even after the learning process due to the randomness in the action choices. Hence, some of the safe \ac{RL} approaches pay special attention to the exploration process, considering some form of risk. 
Unfortunately, it is impossible to avoid undesirable consequences completely without accessing a certain amount of external knowledge (or prior knowledge) of the environment. 
One possible safe exploration strategy is employing Bayesian approaches (e.g. GP~\cite{rasmussen2003gaussian}) for the environment modeling. It can take prior knowledge and predict future trajectories with uncertainties. It allows the agent to explore a region that might result in high reward while maintain constraints by considering the predictive uncertainties~\cite{sui2015safe,wachi2018safe}.  The limitations of these GP based approaches are difficult to extend to high-dimension state space.

The approaches that do not rely on external knowledge, can not avoid catastrophic failures completely, but they attempt to minimise the failures by considering the risk of the exploration. 
For example, Moldovan \& Abbeel~\cite{moldovan2012safe} consider safety using ergodicity, where an action is safe if it is still possible to reach every other state after having taken that action. These methods are limited to small, discrete MDPs where exact planning is straightforward.
For the algorithms can be applied to complex and high-dimensional tasks, Wachi et al. propose MASE~\cite{wachi2023safe} -- algorithm employs an uncertainty quantifier for a high-probability guarantee that the safety constraints are not violated and penalises the agent before safety violation, assuming that the agent has access to an “emergency stop” authority -- namely a human intervention.
Han et al.~\cite{han2015learning}, and Eysenbach et al.~\cite{eysenbach2017leave} propose approaches that learn two policies -- one is for maximising the reward (standard \ac{RL} policy), and the other is for bringing the state back to the initial state (reset policy). Then, they use the reset policy to recover from a potentially unsafe state to maintain their safety. These approaches could fail before they learn when it should use the reset policy and the reset policy itself.
Other simpler approach, Gehring and Precup~\cite{gehring2013smart} define the risk as a magnitude of temporal difference error for the value function. It discounts an expected future reward with the risk to discourage taking risky actions.

As we discussed in the previous section, approaches like \ac{TRPO}, \ac{PPO} and \ac{CPO} are based on a theoretical guarantee that each policy update improves the performance. Hence, they can be seen as safe exploration approaches. However, they do not have built-in mechanisms to ensure avoiding catastrophic failures. Furthermore, \ac{PPO} has been found to suffer from a lack of exploration~\cite{wang2019trust}. So, they might need additional mechanisms to ensure safe exploration.

\section{Incorporating additional data and knowledge}
\label{sec02:external_knowledge}
In reinforcement learning, an agent learns from its own interactions with the environment. However, the agent may sometimes have access to additional knowledge sources or data that can help it improve its performance. For example, the agent may use expert demonstrations, human feedback, or prior knowledge about the task or the environment. This section will explore how to incorporate such knowledge or data into reinforcement learning algorithms. 
We contrast this approach with the one we discussed in the prior sections, where the agent only uses data (trajectory) obtained by itself. However, some approaches in the prior sections can also be used here. 
There are various forms of external knowledge -- trajectory dataset, a computer model (simulator) of the environment and a person with knowledge of the environment. Each form of knowledge holds different characteristics and hence requires a different approach to extract and incorporate the information. We discuss each of them in the following subsections.

\subsection{Trajectory dataset}
If the trajectory dataset is obtained from the interactions between a target environment and an expert human (or an algorithm), then we could rely on a \ac{BC} type of approach that simply learns a mapping from a state to an action from the dataset.
If the dataset is obtained from interactions between a target environment and non-expert, we could employ one of offline \ac{RL} approaches~\cite{agarwal2020optimistic}, which learns an optimal policy from the dataset while avoiding actions out of the dataset distribution.
In case of the dataset is sampled from a similar but not target environment, it requires robust \ac{RL} approach, which takes into account the unknown difference between these environments. It learns a policy that performs well in the worst case of given uncertainty in the environment~\cite{morimoto2005robust, iyengar2005robust}. Alternatively, we can utilise meta-learning to acquire a prior for the target environment. An important note in meta-learning is that the learned prior should cover the actual environment~\cite{harrison2018meta,yamagata2022continuous}. Achieving this requires a dataset that encompasses diverse realisations of the environment. If such diversity is lacking, we may need to relax the prior distribution to ensure it adequately covers the true environment.

The quality of a trajectory dataset is a crucial factor. Ideally, an expert demonstration dataset should encompass all states an agent will likely encounter. If the dataset is obtained from non-experts, ideally, it includes all state-action pairs~\cite{yarats2022}. 
Even when a single trajectory does not encompass the entire optimal path, RL algorithms are capable of learning optimal behaviour from these suboptimal trajectories. This capability, known as \textit{stitching}, is a vital attribute of offline RL algorithms~\cite{fu2020d4rl}. Without the \textit{stitching} ability, an RL algorithm would require a dataset that contains the complete optimal trajectory~\cite{yamagata2023qdt}.

\subsection{Simulators}
Computer models can simulate the environment for many applications and can be used to train an RL agent or create a training dataset. Computer models are often more accessible and less risky than the actual environment, as they do not incur any costs or consequences for failing a task. Therefore, they are useful for pre-training the agent before deploying it to the real environment~\cite{breyer2018flexible}. However, a naive approach may not work well due to the discrepancy between the computer model and the real environment. A possible solution to the issue would be a robust RL approach. This problem is known as "sim-to-real" and has been studied extensively~\cite{zhao2020sim,salvato2021crossing}.

\subsection{Human knowledge}
Human knowledge can be a valuable source of guidance for safe reinforcement learning, but it also poses some challenges. How can we obtain human knowledge in a way that is efficient, reliable and scalable~\cite{goecks2020human}? 
We could use different methods of human input, such as demonstration (where humans provide examples of desired behaviour)~\cite{argall2009survey}, feedback (where humans evaluate the agent's actions)~\cite{Griffith2013} or intervention (where humans correct the agent's actions)~\cite{wang2018intervention,saunders2017trial}. 
Owing to the diverse strategies available for incorporating human knowledge, a dedicated section (Sec.~\ref{sec02:human-in-the-loop} human-in-the-loop) is available for further details.

\section{Human-in-the-loop}
\label{sec02:human-in-the-loop}
A human-in-the-loop approach is another possible approach for achieving safe \ac{RL}. The most robust approach of this category would have a mechanism for a human to intervene in the agent action~\cite{li2022efficient, wang2018intervention}. When the human finds that the agent's action violates constraints, the human takes over the system and applies alternative action instead of the agent. This method guarantees zero violation of the constraints assuming the human can always provide the right actions. However, it is hard to scale to complex environments because the human cost would be prohibitively high. Alternatively, some works propose having a machine-learning model for intervening and replacing unsafe actions with safe ones~\cite{saunders2017trial,marta2021human}, yet they still require the model to be reliable.
Other approaches are that humans advise which actions to take or give feedback regarding the actions the agent just took (i.e. right or wrong)~\cite{Griffith2013, Cederborg2015, yamagata2021safe}. The advice or feedback will guide the agent's learning process and help achieve a good policy quickly.

Reinforcement learning with human feedback is a research area that has gained renewed attention in recent years, especially with the application of ChatGPT~\cite{ouyang2022training,casper2023open}, a conversational agent trained with human preferences. This approach addresses the challenge of defining a suitable reward function for complex tasks by learning from the feedback of human evaluators. This line of research is primarily based on \ac{PbRL}~\cite{akrour2011preference,cheng2011preference,wirth2017survey,christiano2017deep,lee2021pebble,rafailov2023direct}, which learns human preference through relative feedback, such as pair-wise comparisons and rankings. They model human feedback with the Bradley-Terry model~\cite{bradley1952rank} for the pair-wise comparisons and Plackett-Luce model~\cite{duncan1959individual,plackett1975analysis} for the rankings.
The Bradley-Terry model is a special case of the Plackett-Luce model, and it was first introduced by Zermero~\cite{zermelo1929calculation} and heavily studied in the years since, particularly following its rediscovery by Bradley and Terry~\cite{bradley1952rank}. It learns a reward function from the human feedback and then train an agent with the learned rewards, or the agent learn a policy directly from the human feedback.

Furthermore, human feedback can help \ac{RL} agents overcome some of their challenges, such as sparse rewards, misaligned objectives or unsafe exploration~\cite{Cederborg2015,knox2009interactively}. It provides additional guidance, correction, or evaluation of their behaviour through feedback. Therefore, RL with human feedback is an important research area for developing safe and robust RL systems that align with human values and preferences.

\section{Related problems and formulations}
\label{sec02:related_topics}

\subsection{Complex reinforcement learning paradigms}
\label{sec02:other_RL_frameworks}
This paper primarily focuses on the standard \ac{RL} setting. However, in this section, we touch on safety and robustness issues on other \ac{RL} settings -- namely, multi-agent \ac{RL} and hierarchical \ac{RL} settings.

\subsubsection{Multi-agent RL}
\Ac{MARL} is a sub-field of \ac{RL} that focuses on investigating the behaviour of multiple learning agents that coexist in a shared environment. Each agent is motivated by either the global rewards or its own rewards, developing interesting behaviours that can be characterized as collaborative \cite{Tan1997MultiAgentRL}. In some environments, these individual rewards may be opposed to other agents' rewards, resulting in complex group dynamics. 

Safe \ac{MARL} works often define a multi-agent version of \ac{CMDP}. We found three types of the definitions. Some works define the multi-agent version of \ac{CMDP} as a tuple $(\mathcal{S},\{\mathcal{A}_j\}_{j\in\mathcal{N}}, r, \{C_j\}_{j\in\mathcal{N}}, \mathcal{P},\mu,\gamma)$. Where $\mathcal{N}=[1,\dots,n]$ is a set of agents, $\mathcal{A}_j$ is the action space for the agent $j$ and $C_j$ is the set of cost functions for the agent $j$. With this formulation, the rewards are common to all agents, while the cost functions and constraints can be different between agents. Hence, the agents try to maximise the common rewards while maintaining each of their constraints. Each agent's policy $\pi_j$ is optimised as the following equations~\cite{gu2021multi,liu2021cmix}.
\begin{equation} \label{ch2_ma-cmdp_objective_1}
\begin{split}
    \arg\max_{\pi_j} &\mathbb{E}_{\tau\sim\pi_j}\left[ \sum_{t=0}^{\infty}\gamma^t r\left(s_t, a_t, s_{t+1}\right)\right], \\
    &s.t.\ J_{c_i}^\pi \leq \epsilon_i \forall i, \text{ where } c_i \in C_j.
\end{split}
\end{equation}

The second formulation define it as $(\mathcal{S},\{\mathcal{A}_j\}_{j\in\mathcal{N}}, \{r_j\}_{j\in\mathcal{N}}, \{C_j\}_{j\in\mathcal{N}}, \mathcal{P},\mu,\gamma)$, where $r_j$ is the reward function for the agent $j$. This formulation assumes different reward functions amongst agents. The agents try to maximise their own rewards and maintain their constraints. So, each agent's policy $\pi_j$ is optimised as the following equations~\cite{elsayed2021safe}.
\begin{equation} \label{ch2_ma-cmdp_objective_2}
\begin{split}
    \arg\max_{\pi_j} &\mathbb{E}_{\tau\sim\pi_j}\left[ \sum_{t=0}^{\infty}\gamma^t r_j\left(s_t, a_t, s_{t+1}\right)\right], \\
    &s.t.\ J_{c_i}^\pi \leq \epsilon_i \forall i, \text{ where } c_i \in C_j.
\end{split}
\end{equation}

 The last definition is for a distributed scenario, and it defines their \ac{CMDP} as $(\mathcal{S},\{\mathcal{A}_j\}_{j\in\mathcal{N}}, \{r_j\}_{j\in\mathcal{N}}, \mathcal{G}, \{C_j\}_{j\in\mathcal{N}}, \mathcal{P},\mu,\gamma)$,  where $\mathcal{G}=(\mathcal{N}, \mathcal{E})$ indicates available communication link between agents. Each agent receives its own rewards and it tries to maximise the average rewards across all agents~\cite{lu2021decentralized}.
 \begin{equation} \label{ch2_ma-cmdp_objective_3}
\begin{split}
    \arg\max_{\mathbf{\pi}} &\mathbb{E}_{\tau\sim\mathbf{\pi}}\left[ \sum_{t=0}^{\infty}\sum_{j=1}^{n}\frac{1}{n}\gamma^t r_j\left(s_t, a_t, s_{t+1}\right)\right], \\
    &s.t.\ J_{c_i}^\pi \leq \epsilon_i \forall i, \text{ where } c_i \in C_j,
\end{split}
\end{equation}
where $\mathbf{\pi}=\{\pi_j\}_{j=1:n}$ is a set of all agents' policy.

Multi-agent constrained policy optimisation (MACPO)~\cite{gu2021multi} is the multi-agent version of \ac{CPO} algorithm~\cite{achiam2017constrained}. It is the first model-free safe \ac{MARL} algorithm and guarantee monotonic improvement in reward, while theoretically satisfying safety constraints. However, it is computationally expensive and the algorithm has some approximations which causes some errors and troubles in practice~\cite{gu2022review}. Multi-agent proximal policy optimisation (MAPPO)~\cite{gu2021multi} is also model free safe \ac{MARL} algorithm which simplify the MACPO algorithm. However, MACPPO does not provide a guarantee for the constraints.
Safe decentralized policy gradient (Safe Dec-PC)~\cite{lu2021decentralized}  is decentralised algorithm for safe \ac{MARL}. It uses the third multi-agent\ac{CMDP} model in above and assumes each agent has communication link to other (neighbour) agents.

\subsubsection{Hierarchical RL}
Hierarchical reinforcement learning (HRL) is a method that decomposes a reinforcement learning problem into a hierarchy of subproblems or tasks. The most common HRL agent structure consists of two layers of hierarchy. The higher layer invokes the lower-level agents by giving them a sub-goal to achieve or selecting a policy from multiple lower-level policies. The higher layer is trained to maximise the rewards from the target environment, while the lower level policies are trained to maximise intrinsic rewards generated by the high-level agent.

There are some works have been done for safe HRL, however they are quite limited. Hierarchical safe reinforcement learning (HiSaRL)~\cite{xiong2022hisarl} is a two-level hierarchy method; the high-level agent generates a safe and efficient path, and the low-level agent ensures runtime safety with a Lyapunov function-based approach. Hierarchical program triggered reinforcement learning (HPRL)~\cite{gangopadhyay2021hierarchical} employs a structured program for the high-level agent, and it triggers one of the low-level policies that are trained for a specific movement (car manoeuvres, i.e. turn left/right). The structured program in the high-level agent defines rule-based safety specifications, and formal verification is used to ensure safety. The authors of \cite{roza2023safe} propose a two-level hierarchical RL algorithm with the high-level agent providing a sub-goal to the low-level agent. The low-level agent is trained to maximise the intrinsic rewards that the high-level agent generates. It also has a safety layer that replaces a potentially risky action generated by the low-level agent with a safe action. The high-level agent is trained to produce effective sub-goals that maximise the rewards and minimise the safety layer intervention. Limitations of this approach are: i) The safety layer only looks one time-step ahead; hence, it cannot effectively intervene for long time horizon constraints. ii) Its safety guarantee is heavily dependent upon the accuracy of the one-step-ahead cost prediction model. It might be hard to assess the reliability of the DNN model.

\subsection{Beyond reinforcement learning}
\label{sec02:related_fields}
In what follows below, we discuss problems in fields that are closely related to the robust and safe \ac{RL} space. Some of these areas are too large to be properly described in here, so we just touch the each of them and highlight the similarities and differences to the robust and safe \ac{RL} approaches mentioned above.

\subsubsection{Control theory}
The relationship between control theory and RL is that both fields share similar goals, i.e. coming up with a good sequence of actions to achieve desired outcomes.
Control theory starts with a known model dynamics (environment's state transitions), while \ac{RL} assume they are unknown and learn them from interacting with the environment. Control theory can provide insights, methods and guarantees for safe and robust RL. Meanwhile, RL can extend the applicability and scalability of control theory to more complex and data-driven scenarios. Although the difference of the assumption about the model dynamics, they are closely related and idea developed in one field often can be applied to the other field e.g. \ac{MPC} was originally proposed and developed in the control theory society, is commonly used in \ac{RL} field especially for the model-based \ac{RL} planning approaches. Likewise, Lyapunov functions are widely used in control theory to prove the stability of a system, and they can also be applied to reinforcement learning to provide its stability~\cite{clempner2022lyapunov} and safety~\cite{berkenkamp2017safe,chow2018lyapunov}.

\subsubsection{Transfer learning}
Transfer learning is a technique that aims to improve the learning efficiency of a machine learning model on a target task by transferring the knowledge contained in different but related tasks~\cite{pan2009survey, weiss2016survey, zhuang2020comprehensive}. 
Much work has been done to apply transfer learning for RL~\cite{taylor2009transfer,Barreto2017,Kulkarni2016,Schaul2015a}. 
The main benefit of transfer learning for RL is that it can reduce the dependence on many interactions with the target environment, which may be expensive, scarce, or unsafe to obtain. By exploiting the similarities between tasks, transfer learning can improve the learning efficiency and the quality of the learned policies.

\subsubsection{Meta-learning}
Meta-learning, or learning to learn, is the science of systematically observing how different machine learning approaches perform on a wide range of learning tasks and then learning from this experience, or meta-data, to learn new tasks much faster than otherwise possible~\cite{vilalta2002perspective,vanschoren2018meta, hospedales2021meta}. Meta-learning also benefits safe and robust RL by exploiting knowledge from a wide range of tasks and provides a good starting point~\cite{Finn2017} for the agent or priors to the model parameters~\cite{grant2018recasting,harrison2018meta}. Both can significantly reduce the risk of unsafe explorations.

\subsubsection{Sim-to-real}
Sim-to-real is a research area that investigates how to transfer reinforcement learning (RL) agents from simulated environments to real-world settings~\cite{zhao2020sim,matas2018sim,salvato2021crossing}. This is especially relevant for robotics applications, where RL can enable agents to learn complex and adaptive behaviours but also possess difficulties due to the large amount of data needed to learn. 
Employing the target environment simulator can reduce the cost and risk of training RL agents, but it also introduces a discrepancy between the simulation and the reality, known as the sim-to-real gap. This gap can cause the agent to fail or unpredictable behaviour when deployed in the real world. Sim-to-real techniques aim to develop learning algorithms to produce robust models that can handle the gap and ensure safe and reliable performance.

\section{Ethical considerations of safe and robust RL}
\label{sec02:ethical_issues}
RL poses significant ethical challenges, especially when applied to real-world tasks that involve human or environmental impacts~\cite{neufeld2022enforcing}. For an in-depth review of the ethical implications of artificial intelligence in general, we refer the reader to \cite{Coeckelbergh2020-COEAE,Hagendorff2020-HAGTEO-9} and the references therein. Additionally, \cite{ec2019ethics} and similar efforts, provide guidelines to address such ethical considerations for practical applications. In this section, we will focus specifically in some of the main ethical risks that arise from safe and robust RL applications, such as safe rewards, accountability and transparency. 
\paragraph{Reward misspecification} \ac{RL} agents learn a policy that maximises the expected sum of the future rewards. So, if the reward function does not match the true objective of the task (reward misspecification), then the learned policy can be useless or even harmful for people around the agent or the environment (reward hacking~\cite{pan2022effects}). For example, an RL agent that controls a self-driving car may learn to drive recklessly if the reward function only considers speed and not safety. Therefore, designing reward functions that align with the desired outcomes and values of the stakeholders is a crucial ethical challenge in RL.
To mitigate the risk, we can consider \ac{CMDP} framework to properly model various constraints of the task -- if the task is complex and needs to consider various conditions to maintain its safety and fairness. The challenges for \ac{CMDP} are that it often requires high computational power to train the agent, and also, it is hard to guarantee to satisfy all the constraints all the time. 
\ac{CMDP} requires humans to specify all the constraints and rewards correctly, and it is hard to specify all the constraints for some applications.
To mitigate the risk, we could rely on human feedback to estimate the reward function. However, human feedback can be sparse and inaccurate. So, it is required to make the algorithm robust against the sparsity and inaccuracy of the feedback. 

\paragraph{Transparency and accountability} Because \ac{RL} tasks involve a sequence of multiple decisions for the agent's action, it is more difficult to explain the reason behind the decisions than in standard machine learning settings. Also, \ac{RL} agents learn from their own interactions with the environment. It means that the agent's performance could vary depending on its previous policy. That makes it difficult to guarantee its performance and provide accountability to the agent~\cite{milani2023explainable,krajna2022explainability}. 
One possible workaround for the issues is to use the RL agent as a decision support system. In this case, the RL agent provides suggestions for the human action, and then the human makes a final decision about which action to take~\cite{yamagata2020}. The human is accountable for the decision. To make such a system work, the RL agent must provide a reason behind its suggestions so that the human makes a good final decision. The limitation of this approach is that, for many applications, it is not feasible to accommodate human intervention. 

% \paragraph{Safety in the exploration: } \ac{RL} agent explores the environment to find the best behaviour. However, the exploration can cause catastrophic failures or violations of the constraints. Safe exploration is a very important topic for the RL real-world application, and it attracts attention. However, it is still hard to guarantee safety in the exploration of complex, high-dimensional environments. To mitigate the issue, we could rely on external sources of information -- such as human feedback and offline data.

Overall, the ethical issues of \ac{RL} has two aspects -- a risk of misspecified reward and lack of transparency (explainability) of the agent. Safe and robust \ac{RL} can mitigate the first issue, but not so much for the latter one directly. However, the uncertainty estimation algorithms, that is a important component of safe \ac{RL}, helps the transparency of the agent. So, safe and robust \ac{RL} approaches should be able to help making ethical algorithms. However, it is still requires attention to these ethical issues.

\section{A checklist for safe and robust RL}
\label{sec02:check_list}
This section provides practitioners with a list of items and actions to check in order to design safe and robust RL agents. Some of these elements are general principles that apply to any RL problem with particular implications around safety and robustness, while others are specific to certain domains or scenarios. These are organised in four parts, namely, \texttt{specification}, \texttt{additional sources of information}, \texttt{optimisation} and \texttt{safety}. We expect practitioners to navigate through the list on a sequential manner. 

\paragraph{Specification} First of all, we need to clarify the requirements of the task at hand. In addition to the normal RL setting specifications, we should consider the following items:
\begin{itemize} %[noitemsep,leftmargin=.2in]
    \item[S1)] \textbf{Ethical requirements. }For real-world applications, it is important to comply ethical requirements which is often specific to a target application. As the first step in a specification stage, we must list up all ethical requirements. They might be related to its transparency (explainability), its fairness (no discriminatory behaviour) or its privacy (protecting user privacy).
    
    \item[S2)] \textbf{Reward function. }If the reward (or cost) function is defined, ensure it is well aligned with the true objective of the task, and there is no space of \textit{reward hacking}. If the reward function is hard to define, consider relying on human feedback approaches discussed in Sec.~\ref{sec02:human-in-the-loop}.

    \item [S3)] \textbf{Variations. }If the environment can potentially be non-stationary over the foreseen deployment period, the agent must be robust against such variations. We need to specify how much variation could happen (specify uncertainty set) and make the agent robust against the variations with approaches discussed in Sec.~\ref{sec02:optimisation_methods} for robust RL criteria. 

    \item [S4)] \textbf{Constraints. }If there are any constraints the task needs to maintain, we need to specify them explicitly and persist them rigorously. Several forms of constraints are explained in Sec.~\ref{sec02:optimisation_criteria} Constrained RL criterion. Then, the agent must have a method in place to satisfy the constraints. 
\end{itemize}

\paragraph{Additional sources of information: }
Next, we consider what kind of additional knowledge we can exploit. As it is always challenging to maintain safety at the early stage of training, we must exploit any knowledge available prior to the agent interacting with the environment.
\begin{itemize} %[noitemsep,leftmargin=.2in]
    \item [A1)] \textbf{Data.} If there are any trajectory data available, we could exploit them with offline RL, meta-learning or simple \ac{BC} algorithm to learn an initial policy, and then we train it online further. If the trajectory data is obtained from interactions between an expert and the target environment, \ac{BC} would be fine. However, it is not from an expert, we use offline RL algorithm to recover the best policy within the data distribution. If it is not from the actual target environment but from many environments containing the target environment, then we could rely on a meta-learning approach to learn the distribution of environment parameters.

    \item [A2)] \textbf{Simulator. }If there is a computer simulator for the target environment, we can rely on it to train the agent. However, it is important to note that there are always some differences between the simulator and the real world. We can employ sim-to-real algorithms (touched in Sec.~\ref{sec02:related_fields}) or robust RL approaches (Sec.~\ref{sec02:optimisation_criteria}) to overcome the gap. 
 
    \item [A3)] \textbf{Expert. }If there is anyone who knows the task well, obtaining demonstration data or annotations on existing trajectory data is worth considering. Although extracting helpful information from a human is difficult, such data from experts is valuable, especially when the reward is sparse. The expert data directly gives information about the best action at every time step. With the demonstration data, we can rely on offline RL or \ac{BC} algorithm to obtain the initial policy. With the expert annotations on the existing dataset, we can rely on a human feedback algorithm to extract the expert's policy.

    \item [A4)] \textbf{Non-experts. }Obtaining demonstration data or annotations on existing trajectory data from many non-experts is very useful. It is not as data efficient as the expert's data. However, it is possible to work out the best policy from it. Again, we can use the offline RL approach with the demonstration data and one of the human feedback approaches with the annotations. They could work out good policy from the data with mixed quality.
\end{itemize}

\paragraph{Optimisation criteria/method}
We now consider algorithms for optimising/training the agent (policy). We consider three aspects below. They are not mutually exclusive; you likely need to guarantee these three aspects.
\begin{itemize} %[noitemsep,leftmargin=.2in]
    \item[O1)] \textbf{Robustness. }To maintain a certain level of robustness, we need to employ approaches introduced in Sec.~\ref{sec02:optimisation_methods} For robust RL criterion. Under given uncertainties of the environment, these approaches maximise the minimum (the worst case) rewards. 

    \item[O2)] \textbf{Safety. }For keeping the specified constraints, several approaches are discussed in Sec.~\ref{sec02:optimisation_methods}. For constrained RL criterion. Some of them are strong theoretical justifications but computationally expensive. Others are relatively simple algorithms but only empirically proven. 

    \item[O3)] \textbf{Exploration. }Exploring while maintaining safety (constraints) is probably the most challenging objective in RL tasks. Exploring tries something unknown (uncertain); hence, it always has a risk of failure. However, there are some approaches for safe exploration introduced in Sec.~\ref{sec02:sefe_exploration}. 
\end{itemize}

\paragraph{Safety layer}
Finally, we consider applying an extra safety mechanism (safety layer) to guarantee/improve the safety and robustness. The possible approaches are:
\begin{itemize} %[noitemsep,leftmargin=.2in]
    \item[L1)] \textbf{Human intervention. } If the task is feasible to have a human intervention to prevent a failure, this would be the best mechanism to guarantee safety (Sec.~\ref{sec02:human-in-the-loop}). Still, it is important to reduce the number of human interventions by employing some of the abovementioned approaches. Also, it is crucial to show the current status of the environment in an easy-to-understand way so that the person can decide when to intervene. 

    \item[L2)] \textbf{Shielding (formal verification). } Suppose the constraint violation can be detected from the current state of the environment and the agent's action, and there is an action (or sequence of actions) to recover from such a possibly dangerous state. In that case, we can implement a safety layer that detects the constraint violation and replace the action with recovery action.

    \item[L3)] \textbf{Shielding (adaptive). } If it is difficult to have the safety layers mentioned above, it is still possible to have a safety layer that learns when it should intervene and what action to take to recover from the potentially risky states. These approaches cannot guarantee safety, especially during learning. However, it could improve safety if it can learn from a trajectory dataset offline.

    \item[L4)] \textbf{Traceability and explainability. }When the agent fails a task, it is essential to understand why it fails. Such \textit{post-mortem examination} will help understand the failure mechanism and improve the agent algorithm to prevent similar failures in the future. Therefore, traceability (how the agent failed the task) and explainability (why the agent took actions that led to the failure) are important. They do not prevent failures in the current task, but they are essential for preventing similar failures in the future. 

    The basic level of traceability can be achieved simply by recording all trajectories (state, action and reward for every time step). However, further information on the agent's internal states might be required to understand the reason for the agent's action choices. These internal states are also required for explainability. Explainability can be achieved by tracing internal states combined with a mechanism to provide a human-understandable explanation. The mechanism to generate a human-understandable explanation can be challenging, especially when the agent utilises \ac{DNN}, and itself is still a significant research area. 
\end{itemize}

\section{Conclusion}
\label{sec02:summary}
This paper explored the various aspects of safe and robust reinforcement learning (RL), delving into algorithmic frameworks, ethical implications, and practical considerations. The domain of safe and robust reinforcement learning is extensive and multifaceted, covering all relevant literature would be far beyond the scope of any single review.
Our aim was to illustrate various dimensions of this vast field, providing a foundational understanding upon which readers can build. By categorizing existing safe RL algorithms, we have provided a structured overview that summarises the current state of this field. Our aspiration is that this work serves as a resource for a diverse audience, ranging from researchers new to safe and robust RL seeking to understand the overall structure of the field to practitioners aiming to implement safe and robust RL systems in real-world scenarios.

% ========================================================================================
% Chapter 2 END
% ========================================================================================

% \import{}{literature_review.tex}

\begin{appendices}

\section{Summary of literature and the timeline}
\label{literature_summary}

\begin{figure}
	\centering
	\includegraphics[width=1.4\textwidth, angle=90]{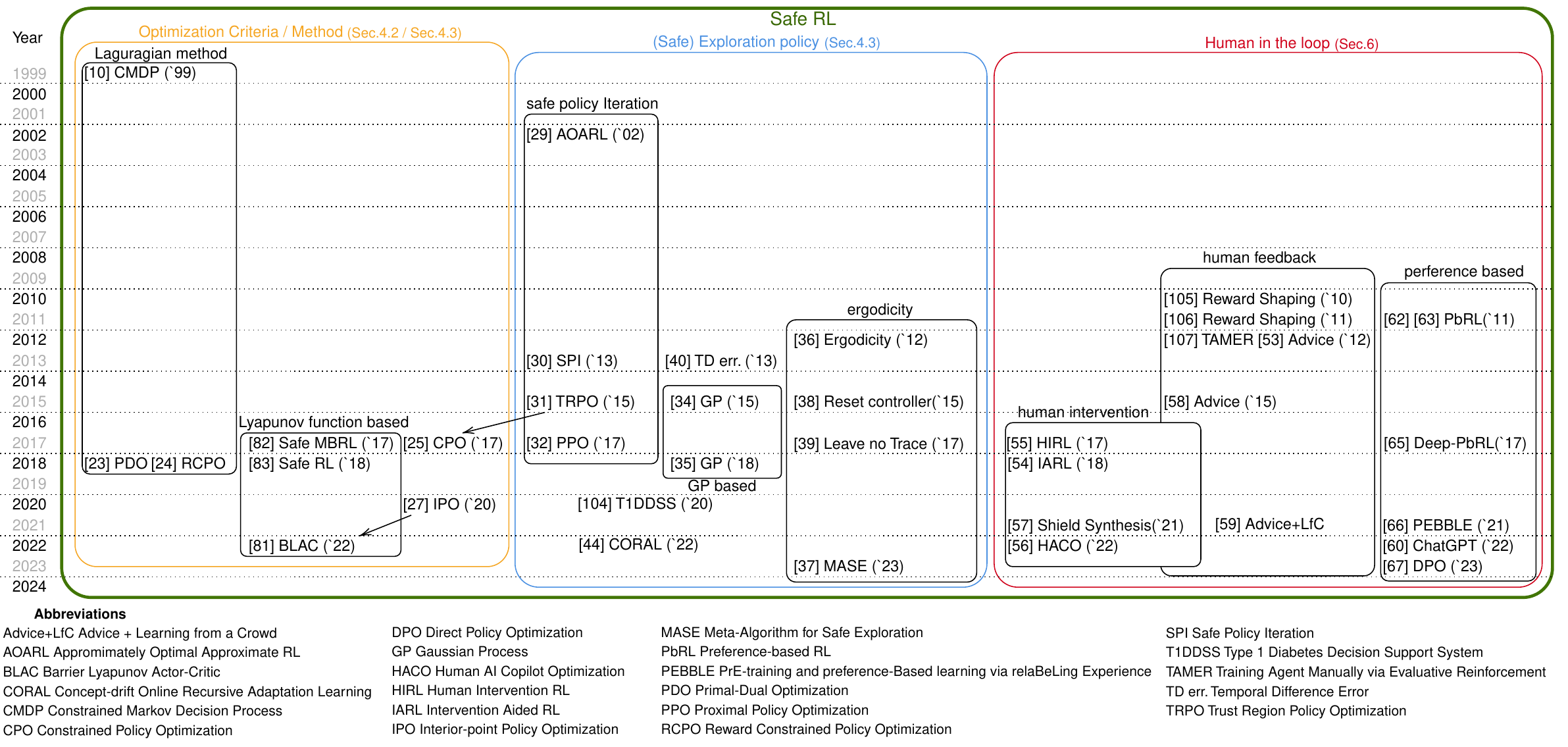}
	\caption[Safe RL literature summary]{The figure organises safe reinforcement learning literature in chronological order, highlighted in colour-coded boxes for major categories and black boxes for sub-categories. Black arrows show inter-paper inspiration across categories.}
	\label{fig02:literature_safe}
\end{figure}

\begin{figure}
	\centering
	\includegraphics[width=1.4\textwidth, angle=90]{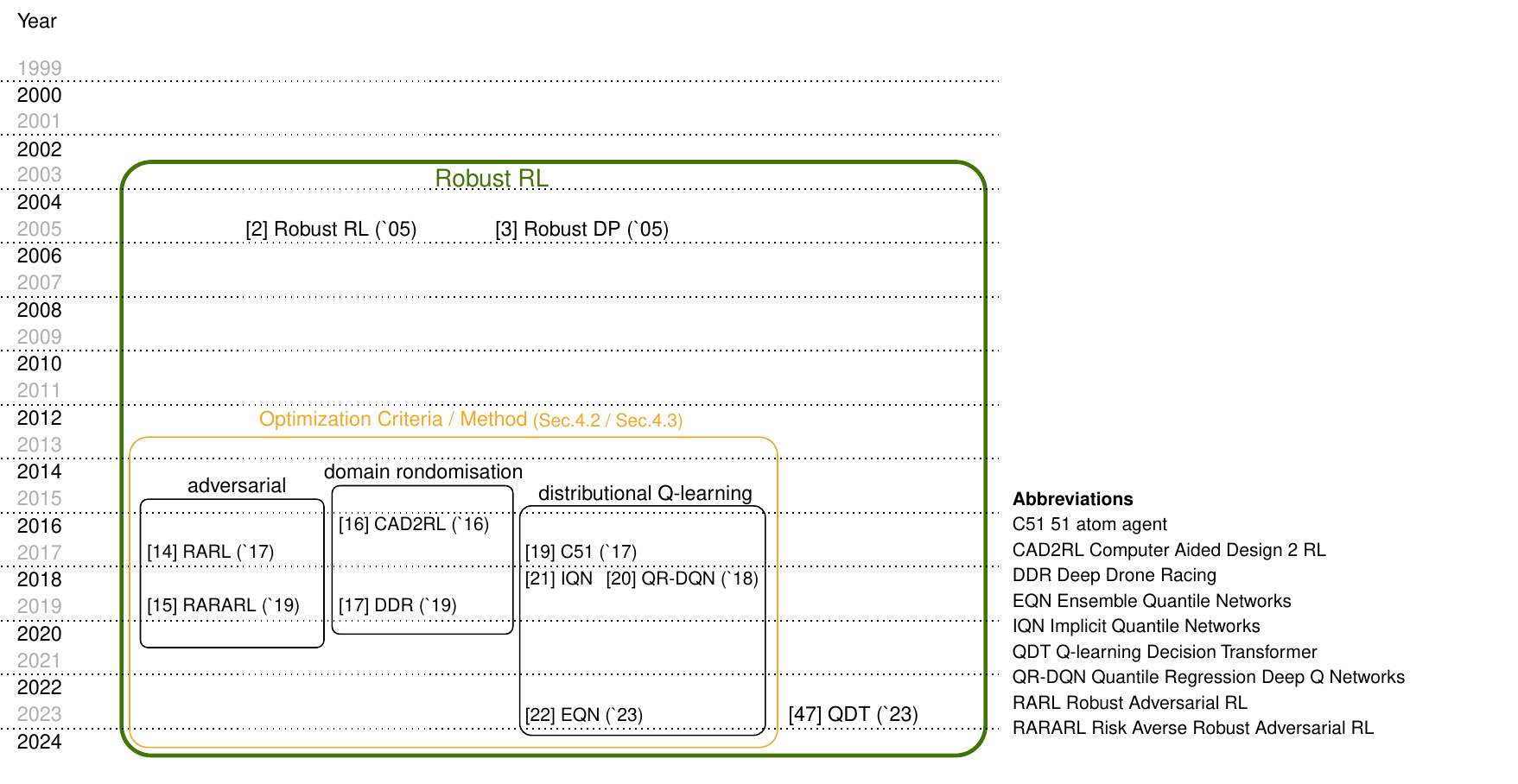}
	\caption[Robust RL literature summary]{The figure organises robust reinforcement learning literature in chronological order, highlighted in colour-coded boxes for major categories and black boxes for sub-categories.}
	\label{fig02:literature_robust}
\end{figure}

Figure~\ref{fig02:literature_safe} and \ref{fig02:literature_robust} show a summary of the safe \ac{RL} and robust \ac{RL} literature, respectively. They categorise the literature into groups and place it in chronological order (from top to bottom). Major categories are highlighted in colour-coded boxes with the relevant sections of this paper, while sub-categories are denoted by black boxes. The black arrows indicate that the paper inspires the other paper in a different category. It is important to note that this summary is not comprehensive; it focuses primarily on significant, recent contributions to the field.

\end{appendices}

\subsubsection*{Acknowledgments}
This work is supported by the UKRI Turing AI Fellowship EP/V024817/1.

\backmatter
\bibliography{thesisbiblio}

\import{}{acronym.tex}

\end{document}

%% file: acronym.tex
% Reinforcement Learning Related
\acrodef{RL}{reinforcement learning}
\acrodef{DRL}{deep reinforcement learning}
\acrodef{DQN}{deep Q-learning}
\acrodef{UDRL}{upside down reinforcement learning}
\acrodef{DT}{decision transformer}
\acrodef{CQL}{conservative Q learning}
\acrodef{IQL}{implicit Q learning}
\acrodef{QDT}{Q-learning Decision Transformer}
\acrodef{BC}{behaviour cloning}
\acrodef{MDP}{Markov decision process}
\acrodef{RMDP}{robust Markov decision process}
\acrodef{CMDP}{constrained Markov decision process}
\acrodef{RTG}{return-to-go}
\acrodef{IPO}{interior point optimisation}
\acrodef{MPC}{model predictive control}
\acrodef{PbRL}{preference-based reinforcement learning}
\acrodef{TRPO}{trust region optimization}
\acrodef{PPO}{proximal policy optimization}
\acrodef{CPO}{constrained policy optimization}
\acrodef{IQN}{implicit quantile networks}
\acrodef{EQN}{ensemble quantile networks}
\acrodef{MARL}{multi-agent reinforcement learning}

% Machine Learning in general
\acrodef{DNN}{deep neural network}
\acrodef{GRU}{gated recurrent unit}
\acrodef{ESN}{echo state network}
\acrodef{RNN}{recurrent neural network}
\acrodef{SGD}{stochastic gradient descent}
\acrodef{GP}{Gaussian process}
\acrodef{RF}{random frorest}
\acrodef{CORAL}{concept-drift online recursive adaptive learning}
\acrodef{MSE}{mean squared error}
\acrodef{RMSE}{root mean squared error}
\acrodef{RLS}{Recursive Least Square}

\acrodef{HMM}{hidden Markov model}
\acrodef{CRF}{conditional random fields}
\acrodef{MEMM}{maximum entropy Markov model}
\acrodef{ALPaCA}{Adaptive Learning for Probabilistic Connectionist Architecture}
\acrodef{MAML}{model-agnostic meta-learning}
\acrodef{BOCPD}{Bayesian online change point detection}
\acrodef{MOCA}{meta-learning via online change point analysis}
\acrodef{MAP}{maximum a posteriori probability}
\acrodef{EM}{expectation-maximisation}
\acrodef{VI}{variational inference}
\acrodef{MI}{mutual information}
\acrodef{BSC}{binary symmetric channel}

% Application specific
% Diabetes
\acrodef{BGL}{blood glucose level}
\acrodef{CGM}{continuous glucose monitor}
\acrodef{BBController}{Basal-Bolus controller}

% Localisation (SPHERE)
\acrodef{RSS}{received signal strength}
\acrodef{PIR}{passive infrared}
\acrodef{AP}{access point}
\acrodef{GNSS}{global navigation satellite system}